%% file: noiseModelsTurk.tex
\documentclass[12pt]{article} %
\usepackage{graphicx} %
\usepackage{fullpage}
\usepackage[numbers]{natbib}
\usepackage{algorithm}
\usepackage{algorithmic}
\usepackage{hyperref}
\usepackage{multirow} 
\usepackage{bm}
\usepackage{dsfont}
\usepackage{amsmath,amssymb}
\usepackage[lofdepth,lotdepth]{subfig}
\usepackage{color}
\usepackage{soul}
\usepackage{wasysym}
\usepackage{float}
\usepackage{enumitem}

\usepackage{titlesec}
\titleformat*{\section}{\LARGE\sffamily\bfseries}
\titleformat*{\subsection}{\Large\sffamily\bfseries}
\titleformat*{\subsubsection}{\large\sffamily\bfseries}
\titleformat*{\paragraph}{\large\sffamily\bfseries}
\titleformat*{\subparagraph}{\large\sffamily\bfseries}
\usepackage{abstract}
\newcommand{\changetitlesize}[1]{\huge{#1}}
\input{macros.tex}

\def\bepsilon{\bm{\epsilon}}
\def\delqm{\delta_{c(m)}}
\def\delbarqm{\overline{\delta_{c(m)}}}
\def\bl{\bm{\ell}}
\def\sig{\sigma(\alpha_n\beta_m)}
\def\bartaun{\overline{\tau_n}}
\def\wrt{with respect to }
\def\qu{query-URL}

\def\Lab{Q(\balpha,\bbeta)}

\def\condsep{|}
\def\vk{k}
\def\vj{j}
\def\modelreg{Continuous-ML}%
\def\modelmix{Ordinal-discrete-mixture}
\def\modelmulti{Dawid-Skene}
\def\modelglad{GLAD}
\def\modelbin{Ord-Binary}
\def\yandex{Yandex}
\def\trec{TREC}

\title{\changetitlesize{\sffamily\bfseries{Inferring ground truth from multi-annotator ordinal data: \\ a probabilistic approach}}}
\author{
    Balaji Lakshminarayanan\footnote{Part of the work was done while at Yandex Labs.}\\
Gatsby Unit, CSML, University College London\\
\texttt{balaji@gatsby.ucl.ac.uk} \\
\and
Yee Whye Teh\\
Department of Statistics, University of Oxford\\
\texttt{y.w.teh@stats.ox.ac.uk} \\
}

\date{}

\begin{document}

\maketitle

\begin{abstract}
    A popular approach for  large scale data annotation tasks is \textit{crowdsourcing}, wherein each data point is labeled by multiple \textit{noisy} annotators.  We consider the problem of inferring ground truth from noisy \textit{ordinal} labels obtained from multiple annotators of varying and unknown expertise levels. 
Annotation models for ordinal data have been proposed mostly as extensions of their binary/categorical counterparts %
 and have received little attention in the crowdsourcing literature. 
We propose a new model for crowdsourced ordinal data   
 that accounts for instance difficulty as well as annotator expertise, and derive a variational Bayesian inference algorithm for parameter estimation. 
We analyze the ordinal extensions of several state-of-the-art annotator models for binary/categorical labels 
 and evaluate the performance of all the models on %
 two real world datasets containing ordinal query-URL relevance scores, collected through Amazon's Mechanical Turk. 
Our results indicate that  the proposed model performs better or as well as existing state-of-the-art methods and is  more resistant to `spammy' annotators (i.e., annotators who assign labels randomly without actually looking at the instance) than popular baselines such as mean, median, and majority vote which do not account for annotator expertise. 
\end{abstract}

\section{Introduction}
Supervised learning tasks such as classification, regression and ranking require features as well as \textit{ground truth}  labels for the training and
evaluation datasets. Unfortunately, obtaining ground truth labels for large datasets is an expensive endeavor. %
\textit{Crowdsourcing} \cite{howe2006rise} is an attractive solution to this problem. In this approach, 
one typically obtains multiple %
 labels for each training instance  from  annotators of unknown and varying expertise levels. 
 Crowdsourcing marketplaces such as Amazon's Mechanical Turk (AMT)\footnote{\url{https://www.mturk.com}}
enable us to collect  labels for large datasets in a time-effective as well as cost-effective manner. The past few years have witnessed a significant increase in the use of crowdsourcing for large scale data annotation tasks in domains such as natural language processing 
\cite{snow2008cheap} and computer vision \cite{sorokin2008utility}.   %

Naturally, the next question is:
how do we handle multiple labels for each training and evaluation instance during the supervised learning process? 
One simple %
approach is to %
estimate the ground truth (for instance, a  weighted combination of the multiple labels) and use this as input to the supervised learning algorithm.  %
Although frequently studied as part of a supervised learning problem, the task of estimating the ground truth from multiple annotations is an interesting problem on its own. 
For example, Dawid et al. \cite{dawid1979maximum} discuss the task of estimating the true response of a patient from patient records,  while Smyth et al. \cite{smyth1995inferring} discuss the task of detecting small volcanoes in Magellan SAR images of Venus.

The critical question is then: how do we optimally combine labels from multiple annotators to form the estimate of the ground truth?
Some simple heuristics for combining the labels are majority vote (mode), mean and median.
However these do not model the fact that annotators can have varying expertise levels, that training instances themselves can have varying difficulties, as well as other characteristics of crowdsourced data.  

In this paper, we develop a %
probabilistic model of the data annotation process, and use Bayesian inference to estimate the ground truth labels.  
The probabilistic modeling approach is very flexible, and a variety of complexities in the annotation process can be incorporated.  
The probabilistic modeling approach can be used to jointly estimate ground truth labels and optimize the parameters of the supervised learning algorithm; cf.~\cite{yan2010modeling, raykar2010learning}; however we do not pursue this approach in this paper and leave it for future work. 
Existing approaches \cite{
    bachrach2012grade,
carpenter2008multilevel,
dawid1979maximum,
karger2011iterative,
raykar2009supervised,
raykar2010learning,
raykarannotation,
rogers2010semi,
snow2008cheap,
welinder2010multidimensional,
whitehill2009whose,
yan2010modeling}
can be broadly categorized according to the following criteria: 
\begin{itemize}
  \item 
  Are the observed and ground truth labels binary/real/ordinal/categorical?
  \item 
  Are ground truth labels (for a subset of training instances) required for training?
  \item 
  Are annotator expertise levels modeled?
  \item 
  Are instance difficulties and/or instance features modeled?
\end{itemize}
One could use combinations of the ideas discussed above as well; for instance, there has been some work on joint modeling of annotator expertise and instance features. 
The above list is not comprehensive; for example, there are differences in the type of parameter estimation technique employed (optimization, expectation maximization, Bayesian inference, etc.) as well.

In this work, we propose a probabilistic model for crowdsourced ordinal annotations.   
Ordinal labels arise naturally in many real world datasets, for example, movie/restaurant ratings and query-URL relevance in information retrieval.  
Annotation models for ordinal data have been proposed mostly as extensions of their binary/categorical counterparts \cite{whitehill2009whose,dawid1979maximum,raykar2009supervised}, which loses the natural ordering of label values and, to the best of our knowledge, have neither been studied in detail nor evaluated experimentally. 
The proposed model models ordinal labels in a natural manner preserving the ordered nature of the label values and 
is in contrast to
most prior work in this area, which have focused on binary, categorical or real-valued labels.
In \cite{raykarannotation}, the work most similar to ours, even though the observed ratings are assumed to be ordinal, the ground truth labels are 
assumed to be binary.  
Real world annotation tasks often involve instances of varying difficulty levels and annotators of varying expertise levels. Crowdsourcing marketplaces typically attract `spammy' annotators, defined as low quality annotators who randomly guess the label without actually looking at the instance, and hence it is necessary to identify and weed out spammy annotators. 
Our model can account for varying levels of instance difficulties, which is useful for active learning (e.g., we could obtain more labels for the difficult instances). Our model can account for varying levels of annotator expertise and in addition, it explicitly models spammy annotators. Hence, our model is able to down-weight spammy annotators and effectively combine the labels from different annotators according to their expertise levels. 
We assume that instance features are not available. 
While some authors have suggested modeling instance difficulty using instance features, it is non-trivial in general to derive features that reflect instance difficulty.  For instance, it is not obvious what characteristics of an image determine the difficulty perceived by the annotators. 
We assume that the ground truth labels are not available for training, which is typically a realistic assumption. %
Our model is very simple, with an %
    efficient variational Bayesian inference algorithm. %

We show that our model subsumes a number of existing models and outperforms popular baselines such as majority vote, median and mode. 
In addition, we explore the ordinal extensions of several state-of-the-art annotator models for binary/categorical labels, and systematically evaluate the performances of the different ordinal annotation models on two real world information retrieval datasets. %
We empirically demonstrate that our model outperforms or performs as well as existing approaches and is more resistant to spammy annotators. 

In Section \ref{sec:model}, we describe our model and inference algorithm.  In Section \ref{sec:relatedwork}, we discuss relevant prior work and elaborate on the relationship between our model and some of the prior work.  We report experimental results in Section \ref{sec:experiments} and conclude in Section \ref{sec:conclusion}.

\section{Proposed \modelmix\ model}\label{sec:model}
In this section, we first describe our problem setup and  the nature of the dataset. Next,  we introduce our proposed model and present variational inference updates for parameter estimation. 

\subsection{Problem setup}\label{sec:problem:setup}
We assume that there are $N$ annotators and $M$ instances (e.g.\ images, \qu\ pairs). Let $r_{nm}$ denote the label provided by the $n^{th}$ annotator to the $m^{th}$ instance, and 
$z_{m}$ denote the (unobserved) \textit{ground truth} label for the $m^{th}$ instance. We assume $z_m$ to be real valued and $r_{nm}$  to take on values on an ordinal scale with $K$ different values, $1, 2, \ldots, K$. 
The proposed model can be easily modified to handle ratings on a $\bv=\{v_1,v_2, \ldots, v_K\}$ scale, where $v_1< v_2 < \ldots < v_K$, and $\{v_k\}$ are known. 
Each instance is typically labeled by very few ($\ll N$) annotators;  hence the observations can be visualized as %
 a $M\times N$ sparse matrix $\bR$. %
Let $\bL$  denote the set of indices where the rating is observed, i.e., $\bL = \{(n,m): r_{nm} \textrm{ is observed} \}$.

To provide a concrete example, the \yandex\ dataset used in our experiments contains $M=10K$ instances of \qu\ pairs and a small subset of $N=51$ annotators are required to assign how relevant an URL is for a specific query on a 5-point scale where $5$ represents highest possible relevance and $1$ represents the lowest possible relevance.  A total of $|\bL|=40K$ annotations were collected, from which our problem is to estimate the ground truth relevance rating (which we assume exists) for each of the 10K \qu\ pairs.

We assume that the training instances which have the same difficulty can be grouped into  \textit{categories}. For instance, in the web search example, if we assume all query-URL pairs belonging to the same query are equally difficult, the category could refer to the query corresponding to each \qu\ pair.  
 Let there be $C$ ($\le M$) categories and let $c(m) \in \{1,2,\ldots,C\}$ denote the category of the $m^{th}$ instance. %
In the case where the category is not observed, 
 one might interpret category as a modeling choice that allows us to control the granularity level at which we model instance difficulty (cf. Section~\ref{sec:comparison}); for instance, one could set $c(m)=m$ (every instance is treated separately) or $c(m)=1$ (every instance is treated equally).
 
Let $\bl_{m}$  denote the set of annotators corresponding to the $m^{th}$ instance, $\bl_{n}$ denote the set of instances corresponding to the $n^{th}$ annotator, and let $\bl_{\cdash}=\{(n,m)\in\bL: c(m)=\cdash\}$ denote the set of ratings corresponding to category $\cdash$. 

\begin{figure}%
\centering
\includegraphics[width=0.9\columnwidth]{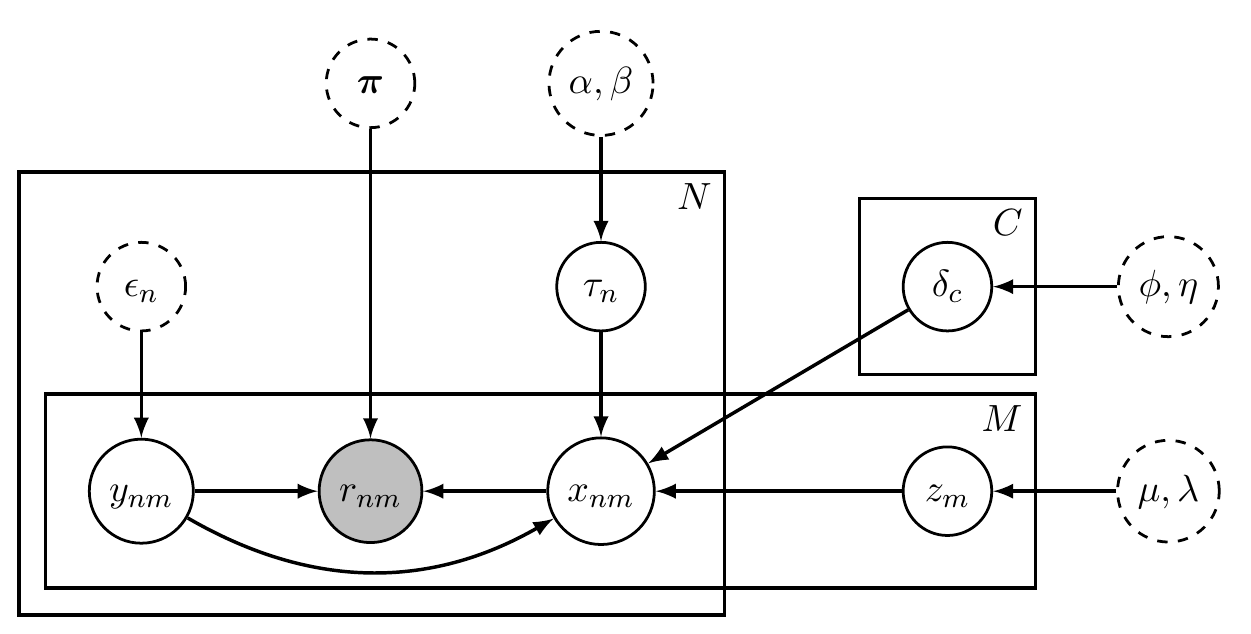}
\caption{Graphical model for \modelmix\ model: $\alpha, \beta, \mu, \lambda, \phi, \eta, \bpi, \{\epsilon_n\}_{n=1}^N$ are   parameters, $\{\delta_{c}\}_{c=1}^C, \{\tau_{n}\}_{n=1}^N$, $\{z_{m}\}_{m=1}^M$, $\{x_{nm}, y_{nm}\}_{nm\in\bL}$ are  latent variables and $\{r_{nm}\}_{nm\in\bL}$ are the observed ratings.}
\label{fig:modelmix}
\end{figure}

\subsection{\modelmix \ model}\label{sec:modelmix}
In this section we describe our proposed graphical model, which is shown in its entirety in Figure~\ref{fig:modelmix}.  
The objective of inference is to estimate the ground truth
value $z_{m}$ for the $m^{th}$ instance.  We assume 
a Gaussian prior for $z_m$ 
with density $\mathcal{N}(\cdot\condsep\mu,\lambda^{-1})$, where the mean is $\mu$ and the precision (inverse variance) is $\lambda$. 
We model the observed rating $r_{nm}$ as a draw from a mixture model with two components, one dependent on the ground truth $z_m$ and a second `spam' component (i.e., independent of $z_m$) that is shared across annotators and instances. We denote the choice between the two components using a binary  random variable $y_{nm}$.  The prior mean of $y_{nm}$, denoted by $\epsilon_n$, is annotator specific, to allow for varying levels of noise in the annotations of different annotators. When $y_{nm}=0$, the rating is drawn from the spam component distribution, which is simply a discrete distribution with probabilities $\bpi$. %
Hence, $1-\epsilon_n$ can be interpreted as the \emph{spamminess} measure of the $n^{th}$ annotator.

When $y_{nm}=1$,  the ordinal-valued $r_{nm}$ is modeled as follows: we first draw $x_{nm}$ from a Gaussian distribution centered at the true value $z_{m}$, then map the continuous-valued $x_{nm}$ to an ordinal $r_{nm}$ deterministically by simply thresholding.  Let
$b_0, b_1, \ldots, b_K$ denote a series of thresholds.  Then $r_{nm}=k$ for the smallest $k$ such that $b_k$ is larger than $x_{nm}$, i.e., $b_{k-1}\leq x_{nm}<b_k$.  
In our experiments, $K=5$, and we simply fix $b_0=0.5,b_1=1.5, b_2=2.5, b_3=3.5, b_4=4.5, b_5=5.5$.
One could learn user specific thresholds as well, but this causes non-identifiability in the $z_m$ values which then may not be on the same scale as $r_{nm}$. %
 See \cite{rogers2010semi} for a related discussion in the context of ordinal regression.
We model the dependence of $x_{nm}$ on the ground truth $z_{m}$ using a Gaussian distribution with annotator and category  specific noise precision.  For simplicity, we take this precision to be $\tau_{n}\delqm$, where the positive latent variable $\tau_n$ can be interpreted as the \text{expertise}  of the $n^{th}$ annotator, i.e., higher 
$\tau_n$ implies lesser variance around the true value $z_m$. %
The positive latent variable $\delta_{c}$ can be interpreted as the 
\emph{inverse difficulty} of 
the $c^{th}$ category.  
If $\delta_{c}<1$, the category is difficult and even annotators whose
 ratings usually have high precision can exhibit lower precision.  Note that $\delta_c$ is shared by all instances corresponding to the $c^{th}$ category.   
 In summary, we have $x_{nm} \sim \mathcal{N}(\cdot\condsep z_m,(\tau_n\delqm)^{-1})$. 
 Finally, we impose  independent gamma priors on $\tau_n$ and $\delqm$.

 The complete generative process is shown in Figure~\ref{fig:modelmix}.  The conditional densities are given below:
\begin{align}
p(\bR| \bY, \bX, \btau,\bz,\bdelta,\bepsilon,\bpi)
&= \prod_{nm \in \bL} \pi_{r_{nm}}^{1-y_{nm}} \Bigl(
 \indicator[b_{r_{nm}-1}\leq x_{nm}< b_{r_{nm}}]
\Bigr)^{y_{nm}}, \label{eq:mixture} \\
 p(\bY|\bepsilon)&=\prod_{nm\in\bL}\bernoulli(y_{nm}\condsep\epsilon_n),  \nn\\
 p(\bX|\bz,\btau,\bdelta,\bY)
 &=\prod_{nm\in\bL}\Bigl(\mathcal{N}(x_{nm}\condsep {z_{m}},(\tau_{n}\delqm)^{-1})\Bigr)^{y_{nm}}, \nn\\
 p(\btau) &= \prod_{n} \mathcal{G}(\tau_{n}\condsep\alpha,\beta), \nn \\
p(\bdelta) &= \prod_{c}\mathcal{G}(\delta_{c}\condsep\phi,\eta), \nn \\
 p(\bz) &= \prod_{m}\mathcal{N}(z_{m}\condsep\mu,\lambda^{-1}), \nn%
\end{align}
where $\indicator[\cdot]$ denotes the indicator function, $\mathcal{G}(\cdot\condsep\alpha,\beta)$ denotes the gamma density with shape parameter $\alpha$ and inverse scale parameter 
$\beta$ and $\bernoulli(\cdot\condsep\epsilon)$ denotes a Bernoulli probability with mean $\epsilon$. %
In the following, the hyperparameters are $\btheta = \{\alpha,\beta,\mu,\lambda, \phi, \eta, \bepsilon, \bpi\}$.  In our experiments, we set all the entries of $\bpi$ to $1/K$, i.e., we assume that the ratings from a spammy annotator are uniformly distributed. %
In our experiments, we learn %
$\alpha,\beta$ and $\bepsilon$ using type II maximum likelihood within the variational Bayesian inference algorithm, while we fix $\phi=10, \eta=5$ to ensure identifiability\footnote{Note that there is a multiplicative degree of freedom since $\tau_n$ and $\delqm$ only appear as the product $\tau_n\delqm$ in the precision term.} 
and fix $\mu$ to the mean of $\bv$ (ordinal scale) and $\lambda=0.1$. %

\subsection{Parameter Estimation}
The marginal likelihood of the observed annotations is given by
\begin{align*}
 p(\bR|\btheta) = \expect_{\btau,\bz,\bdelta,\bX,\bY} \ [p(\bR|\btau,\bz,\bdelta,\bX,\bY)].
\end{align*}
Since both this and the posterior distribution are intractable, we use a variational Bayes (VB) algorithm \cite{Beal03}
for parameter estimation.  Alternatively one may choose to use Markov chain Monte Carlo, but this can be slower to run and collect enough samples to estimate $z_m$ well.
In VB, the log marginal likelihood is lower bounded by the negative variational free energy: 
\begin{align}
\ln p(\bR|\btheta) & =\freeEn(q,\btheta)+\KL{q(\bz,\btau,\bdelta,\bX,\bY)}{p(\bz,\btau,\bdelta,\bX,\bY|\bR,\btheta)}, \nn%
\end{align}
where $p(\bz,\btau,\bdelta,\bX,\bY|\bR,\btheta)$ denotes the true posterior, $q(\bz,\btau,\bdelta,\bX,\bY)$ the approximate variational posterior, and $\freeEn(q,\btheta)$ denotes the variational lower bound (i.e.,\ the negative variational free energy):
\begin{align}
\freeEn(q,\btheta) & =\expect_q[\ln p(\bR,\bz,\btau,\bdelta,\bX,\bY|\btheta)]+\entropy[q]. \nn %
\end{align}
Note that $\expect_q[\cdot]$ denotes expectation \wrt the (variational) distribution $q$ and  $\entropy[q]$ denotes the entropy of  $q$.   
We assume that the variational distribution $q(\btau,\bz,\bdelta,\bX,\bY)$ factorizes as follows:
\begin{align}\label{eq:mf}
q(\btau,\bz,\bdelta,\bX,\bY)=\prod_{n}q(\tau_{n}) \prod_{m} q(z_{m}) \prod_{c} q(\delta_{c})\prod_{nm\in\bL}q(x_{nm}, y_{nm}).
\end{align}
Note that we model the mixture indicator $y_{nm}$ and the continuous-valued latent variable $x_{nm}$ jointly in the variational posterior.  This is because of the deterministic relationship between $x_{nm}$ and $r_{nm}$ when $y_{nm}=1$, which induces strong dependence between $x_{nm}$ and $y_{nm}$ in the true posterior.  Since $x_{nm}$ and $r_{nm}$ are independent when $y_{nm}=0$, it is sufficient to keep track of $q(x_{nm},y_{nm})$ via $q(y_{nm})$ and $q(x_{nm}|y_{nm}=1)$ only.  Furthermore, since $x_{nm}$ cannot lie outside the range of the interval $[b_{r_{nm}-1},b_{r_{nm}})$ when $y_{nm}=1$, we will see that the optimal $q(x_{nm}|y_{nm}=1)$ is simply a truncated Gaussian---its conditional prior distribution given $z_m$ and $\tau_{n}\delqm$ limited to the interval $[b_{r_{nm}-1},b_{r_{nm}})$.
    
    Alternatively, the latent variable $x_{nm}$ can be integrated out in (\ref{eq:mixture}), and $r_{nm}|z_m$ can be expressed as a difference of two Gaussian CDFs as in ordinal regression \cite{chu2005gaussian}.  However treating $\{x_{nm}\}$ as latent variables leads to much simpler variational updates and we have found that it works sufficiently well in practice. 
Such an approach has been  successfully applied for
multinomial probit regression in \cite{girolami2006variational}.

\subsection{Variational updates}
For completeness, we provide the full set of variational updates in this section.  The variational approximation leads naturally to a Gaussian posterior $q(z_{m})$, Bernoulli posterior $q(y_{nm})$,  and gamma posteriors  $q(\tau_{n})$, $q(\delta_{c})$.  As mentioned earlier, $q(x_{nm}|y_{nm}=1)$ has the form of a truncated Gaussian, whose density we denote as $\mathcal{TN}(\cdot\condsep\mu,\sigma^2,l,u)$, where the Gaussian mean is $\mu$, variance is $\sigma^2$, and the lower and upper limits are given by $l$, $u$ respectively.  We parametrize the variational posteriors using variational parameters as follows:
\begin{align}
q(z_{m}) &=  \mathcal{N}(z_{m}|\mu_{m},\lambda_{m}^{-1}), \label{eq:yg:posterior}\\
q(\tau_{n}) &= \mathcal{G}(\tau_{n}|\alpha_{n},\beta_{n}), \nn\\
q(\delta_{c}) &= \mathcal{G}(\delta_{c}|\phi_{c},\eta_{c}), \nn\\
q(y_{nm}) &= \bernoulli(y_{nm}|\omega_{nm}), \nn\\
q(x_{nm}|y_{nm}=1) &= \mathcal{TN}(x_{nm}|\nu_{nm}, \rho_{nm}^{-1}, b_{r_{nm}-1}, b_{r_{nm}}). \nn
\end{align}
The variational E-step updates are:
\begin{align}
\lambda_{m} &\gets \lambda +  \delbarqm\sum_{{n\in\bl_{m}}}\ \overline{y_{nm}}\ \bartaun, %
&\mu_{m} &\gets \frac{\mu\lambda + \delbarqm\sum_{{n\in\bl_{m}}}\ \bartaun\ \overline{y_{nm}}\ \overline{x_{nm}}}{\lambda + \delbarqm\ \sum_{{n\in\bl_{m}}} \overline{y_{nm}}\ \bartaun} ,  \nn \\ %
\alpha_{n} &\gets \alpha + \frac{1}{2}\sum_{m\in\bl_{n}} \overline{y_{nm}} , %
&\beta_{n} &\gets \beta + \frac{1}{2} \sum_{m \in \bl_{n}} \delbarqm\ \overline{y_{nm}}\ \expect_q[(x_{nm}-z_{m})^{2}],
\nonumber\\ 
\phi_{c} &\gets \phi + \frac{1}{2}\sum_{(n,m)\in\bl_c}\overline{y_{nm}}, %
&\eta_{c} &\gets \eta + \frac{1}{2} \sum_{(n,m)\in\bl_{c}} %
\bartaun\ \overline{y_{nm}}\ \expect_q[(x_{nm}-z_{m})^{2}],
\nonumber\\
\nu_{nm} &\gets \overline{z_m},
&\rho_{nm} &\gets \bartaun\ \delbarqm,\nn
\end{align}
\begin{align}
\omega_{nm} &\gets \frac{\epsilon_n\sqrt{\frac{\bartaun\delbarqm}{2\pi}}\exp\Bigl(-\frac{\bartaun\delbarqm}{2} \expect_q[(x_{nm}-z_{m})^{2}]\Bigr)}{\epsilon_n\sqrt{\frac{\bartaun\delbarqm}{2\pi}}\exp\Bigl(-\frac{\bartaun\delbarqm}{2} \expect_q[(x_{nm}-z_{m})^{2}]\Bigr) + (1-\epsilon_n)\prod_k\pi_k^{\indicator[r_{nm}=\vk]}}, 
\end{align}
where 
\begin{align*}
    \expect_q[(x_{nm}-z_{m})^{2}] =\overline{x^{2}_{nm}} -2\overline{x_{nm}}\ \overline{z_{m}}+\overline{z_m^2}.%
\end{align*}
Note that an overbar denotes the expectation of the corresponding variable \wrt %
its variational distribution, i.e., $\overline{z_m}=\mu_m, \overline{z_m^2}=\mu_m^2+1/\lambda_m$, $\overline{y_{nm}}=\omega_{nm}$, $\bartaun=\alpha_n/\beta_n$, $\delbarqm=\phi_c/\eta_c$, and $\overline{x_{nm}}, \overline{x^2_{nm}}$ denote the first two moments of the truncated Gaussian distribution \emph{given $y_{nm}=1$}. 
Maximizing $\freeEn(q,\btheta)$ \wrt$\epsilon_n$, we  obtain
\begin{align}%
    \epsilon_n &\gets \frac{1}{|\bl_n|}\sum_{m\in\bl_n}\overline{y_{nm}}.
\end{align}
The update for $\alpha$ and $\beta$ involves just maximum likelihood estimation for the gamma distribution. %

\subsection{Prediction of ground truth}
The optimal prediction (in the mean-squared sense) is the posterior mean, i.e.,\ $\hat{z}_m=\E[z_{m}|\bR,\btheta]$. We can approximate this using  the mean under the variational posterior defined in (\ref{eq:yg:posterior}), i.e.,\ $\hat{z}_m=\overline{z_m}$.

\section{Related work}\label{sec:relatedwork}
In this section, we provide an overview of existing approaches for dealing with ordinal data and highlight the connections between the proposed model and previous models. 
Rogers et al. \cite{rogers2010semi} proposed a multi-annotator ordinal regression model involving a Gaussian process prior over the function mapping instance features to ground truth label. In this paper, we assume that instance features are not observed and $z_m$ are independent, though in principle, one could assume a GP prior over $\bz$ in the flavor of \cite{rogers2010semi}. 
When $\epsilon_{n}=1, \forall n$ and $\delta_c=1, \forall c$, 
we obtain the model in \cite{rogers2010semi}\footnote{ 
    Note that $\epsilon_n=1, \forall n$ implies that $y_{nm}=1, \forall n,m$, corresponding to the case where the data is always generated according to the ordinal mixture component.}.
Another key difference is that we use variational inference unlike  \cite{rogers2010semi} who  propose a Gibbs sampling algorithm, which leads to significant computational gains and hence our algorithm is scalable to large datasets. 

We also consider models that simplify the ordinal labels to one of the following label types: 
\begin{itemize}[leftmargin=0cm,itemindent=0pt,labelwidth=\itemindent,label={}]
        \item \textbf{\textit{Continuous labels:}} In this case, the observed labels as well as the ground truth labels are assumed to be real-valued. 
    If we remove the ordinal mapping in \modelmix\  model, i.e., set $r_{nm}=x_{nm}$, the model can produce continuous ratings. Raykar et al. \cite{raykar2010learning} suggested the following model for real-valued ratings: $r_{nm}\sim\Normal(r_{nm}|{z_{m}},\tau_{n}^{-1})$. Note that 
    we obtain this model 
    when 
    $\delta_{c}=1,\forall c$, i.e., instance difficulty is not modeled and
    $r_{nm}=x_{nm},\forall n,m, \epsilon_{n}=1,\forall n$. Another key difference between \modelmix\ model and \cite{raykar2010learning} is that we impose gamma priors on $\btau$ and use a variational inference algorithm rather than a maximum likelihood solution (which can lead to arbitrarily large values of $\btau$). We refer to this model %
    as the \emph{\modelreg}\ model.%
\item \textbf{\textit{Multi-class labels:}} In this case, the observed labels as well as the ground truth labels are treated as (discrete) class labels and the relative order of the labels is ignored.
We consider the following two models for multi-class labels: 
\begin{itemize}[leftmargin=0cm,itemindent=0pt,labelwidth=\itemindent,label={}]
\item \emph{\modelmulti}: this model was  proposed by \cite{dawid1979maximum}. This model uses $O(K^2)$ parameters per annotator and does not account for instance difficulty.  Further details are available in Appendix~\ref{sec:modelmulti}. 
\item  \emph{\modelglad}:  this is the multi-class extension proposed in \cite{whitehill2009whose}. This model uses $O(K)$ parameters per annotator and accounts for instance difficulty. Further details are available in Appendix~\ref{sec:glad}.
 \end{itemize}
\item \textbf{\textit{Binary labels:}} Raykar et al. \cite{raykar2009supervised} suggested  an extension of their binary noise model to  ordinal labels by reducing the  ordinal labels to $K-1$ binary variables by defining $\tilde{z}_{mk}= \indicator[z_m>k], \tilde{r}_{nmk}= \indicator[r_{nm}>k], 1\leq k < K$ \cite{frank2001simple}. %
    In this case, the observed labels as well as the ground truth labels are treated as ordinal labels. Note that the ground truth labels are assumed to be real-valued in our model. 
 While other annotator models for binary labels have been explored in \cite{raykar2009supervised,whitehill2009whose,carpenter2008multilevel,yan2010modeling,welinder2010multidimensional,karger2011iterative}, for simplicity, we restrict our attention to the binary noise model proposed in \cite{raykar2009supervised}. We refer to this model as the \emph{\modelbin} model. %
 Further details are available in Appendix~\ref{sec:modelbin}. 
\end{itemize}
Our proposed \emph{\modelmix}\ model as well as the \emph{\modelreg}\ model are applicable only in scenarios where the relative differences in the ordinal scale $\{v_1,v_2,\ldots,v_K\}$ can be quantified. However, the \emph{\modelmulti}, \emph{\modelglad} and \emph{\modelbin} models are applicable even when the relative differences are not quantified (for instance, $\{v_1,v_2,v_3\}=$\{cold, warm, hot\}). 

Note that none of the models discussed above contain a mixture component for handling spammy ratings. 
 To the best of our knowledge, the use of a mixture component for handling spammy ratings is novel. %
 A two-component mixture model was proposed in \cite{bachrach2012grade} for modeling students' responses. If the student knows the correct answer, the observed response is the same as the ground truth (i.e., there is no noise model), else the observed response is generated from a noise distribution. %
 However, neither the observed ratings nor the ground truth ratings are ordinal in \cite{bachrach2012grade}. %

\section{Experimental results}\label{sec:experiments}
\subsection{Dataset}
We evaluate the models on two datasets, namely the \yandex\ and \trec\ datasets. The \yandex\ dataset %
 consists of 40,340 ratings corresponding to 51 annotators, 601 queries and  10,462 \qu\ instances collected through Amazon's Mechanical Turk. Hence, there are about 17.4 URLs per query and 3.85 ratings per \qu\ instance on average. The annotators were shown a \qu\ pair and asked to rate the relevance of the URL for that particular query on a 5-point scale, with 5 representing highest possible relevance and
1 representing lowest possible relevance. 
 The ground truth ratings are available for all the %
10,462 \qu\ pairs and were collected from in-house 
 expert annotators. %
The original \trec\ dataset used in \cite{Buckley10-notebook} consists of 98,453 ratings corresponding to 766 annotators, 
100 queries and 20,232 query-URL instances. The ground truth ratings are available for only 3277 instances out of the the 20,232 query-URL instances. %
We slightly processed this dataset\footnote{
    The ratings were originally on a $\{-2,-1,0,1,2\}$ scale where $-1$ corresponds to missing ground truth label and $-2$ corresponds to a broken link. We excluded the annotator ratings with value $-2$, and mapped the values from $\{0,1,2\}$ to $\{1,2,3\}$. This mapping affects the value of NDCG, but does not affect the values of MSE and correlation.}
to obtain a dataset containing 
91,783 ratings on a 3-point scale, 
corresponding to 762 annotators, 
100 queries and 20,026 query-URL instances. 
\subsection{Evaluation}
We use the following performance metrics to compare the methods:
\begin{itemize}
\item \emph{Mean squared error (MSE)}: $\frac{1}{M}\sum_{m} (z_{m} - \hat{z}_{m})^{2}$,
\item \emph{Pearson's correlation coefficient (Correlation)}: $\frac{1}{M-1} \frac{\sum_{m} 
  (z_{m}-\bar{z})( {\hat{z}}_{m} - \bar{\hat{z}})}{\sqrt{\Var(z)\Var(\hat{z})}}$, 
  where $\Var(Z)$ denotes variance of Z. 
 \item \emph{Normalized Discounted Cumulative Gain (NDCG)}:  
     When the estimated ground truth values are used to train or evaluate a supervised ranking algorithm, we 
     care about the relative differences between the ground truth estimates of URLs corresponding to the same query 
     rather than the absolute values of the ground truth estimates. 
     NDCG is a ranking measure that evaluates how well a list of URLs is ranked compared to the ideal ordering (i.e., URLs sorted in desending order of ground truth relevance values) \cite{liu2009learning}. 
Note that NDCG is a query level metric unlike MSE %
and Correlation, which are \qu\ level metrics. %
\end{itemize}
Lower values of MSE, and higher values of correlation and NDCG indicate better performance. 

\subsection{Simulation details}
The inference algorithms might converge to a local optimum. Hence, for each model, we initialize the inference algorithm 10 times with different initializations
and compute the performance metrics using the predictions corresponding to the parameter settings with the highest (variational) lower bound. We restrict the maximum 
number of iterations to 1000 and stop if the absolute difference between the lower bounds, $\Delta\freeEn(q,\btheta)$, is less than $0.1$. We implemented all our scripts in MATLAB. The scripts can be downloaded from the authors' webpages.

\subsection{Comparison of the methods}\label{sec:comparison}
We consider the \modelmix \  model with three different configurations:
\textit{\modelmix\ (\qu)}, where each instance is treated as a separate category $(c(m)=m)$, \textit{\modelmix\ (query)}, where all URLs for a given query belong to the same category, and 
with a slight abuse of notation, %
\textit{\modelmix},
where instance difficulty is not modeled ($c(m)=1$). We compare the \modelmix \ model to other models described in Section \ref{sec:relatedwork} as well as simple baselines such mean, median and majority-vote\footnote{Note that we assumed the ordinal scale $\{v_1,v_2,\ldots,v_K\}$ is known; hence, it is possible to compute the mean and median in a meaningful way.}. The results are shown in Table~\ref{tab:comparison}, with the proposed model variants highlighted in bold. We observe that in terms of all the metrics, the \modelmix\ model performs better than majority-vote, mean and median. 
Perhaps surprisingly, modeling instance difficulty does not seem to improve performance in the \yandex\ dataset and all three variants of \modelmix\ model perform similarly. However, modeling instance difficulty improves performance in the \trec\ dataset and the \textit{\modelmix\ (\qu)} variant performs the best. 
Amongst the models discussed in Section \ref{sec:relatedwork}, the \emph{\modelmulti}\ model achieves the best overall performance on both the datasets (although the \emph{\modelreg}\ model, not surprisingly, achieves the lowest MSE), suggesting that reducing ordinal
labels to multi-class labels is better than reducing them to continuous or binary labels. 
Amongst the multi-class label methods (see Section \ref{sec:relatedwork}), the \emph{\modelmulti} model outperforms the \emph{\modelglad} model indicating that modeling the full confusion matrix is beneficial.

\begin{table}[htdp]
\centering
\subfloat[][\yandex]{
    \begin{tabular}{|l|c|c|c|}
        \hline
        &\textit{MSE $\downarrow$}&\textit{Correlation $\uparrow$}&\textit{NDCG $\uparrow$}\\\hline
  \textit{Dawid-Skene}&0.706&0.663&0.951\\\hline
        \textit{GLAD (query-URL)}&0.760&0.614&0.945\\\hline
        \textit{Ord-Binary}&0.914&0.644&0.950\\\hline
        \textit{Ord-Continuous-ML}&0.691&0.664&0.952\\\hline
        \textit{Majority-Vote}&0.866&0.604&0.936\\\hline
        \textit{Mean}&0.699&0.664&0.951\\\hline
        \textit{Median}&0.775&0.638&0.942\\\hline
        \textbf{\textit{Ordinal-discrete-mixture }}&0.653&0.667&0.951\\\hline
        \textbf{\textit{Ordinal-discrete-mixture  (query)}}&0.653&0.668&0.951\\\hline
        \textbf{\textit{Ordinal-discrete-mixture  (query-URL)}}&0.654&0.667&0.951\\\hline
    \end{tabular}
\label{tab:comparison:yandex}
}
\\
\subfloat[][\trec]{
\begin{tabular}{|l|c|c|c|}
\hline
   &\textit{MSE $\downarrow$}&\textit{Correlation $\uparrow$}&\textit{NDCG $\uparrow$}\\\hline
\textit{Dawid-Skene}&0.698&0.423&0.923\\\hline
\textit{GLAD (query-URL)}&0.736&-0.045&0.854\\\hline
\textit{Ord-Binary}&0.893&0.422&0.919\\\hline
\textit{Ord-Continuous-ML}&0.685&0.278&0.901\\\hline
\textit{Majority-Vote}&0.854&0.318&0.905\\\hline
\textit{Mean}&0.649&0.359&0.911\\\hline
\textit{Median}&0.746&0.336&0.907\\\hline
\textbf{\textit{Ordinal-discrete-mixture }}&0.675&0.318&0.908\\\hline
\textbf{\textit{Ordinal-discrete-mixture  (query)}}&0.616&0.399&0.919\\\hline
\textbf{\textit{Ordinal-discrete-mixture  (query-URL)}}&0.606&0.413&0.921\\\hline
\end{tabular}
\label{tab:comparison:trec}
}
\caption{Comparison between different methods on \yandex\ and \trec\ datasets: For models that account for instance difficulty, the granularity is shown in parenthesis. The proposed \modelmix\ model (highlighted in bold) outperforms or performs as well as existing state-of-the-art methods.} %
\label{tab:comparison}
\end{table}

\subsection{Effect of \emph{spammy} ratings}\label{sec:expt:spam}
\begin{figure}[htbp]
\begin{center}
\includegraphics[scale=1.0]{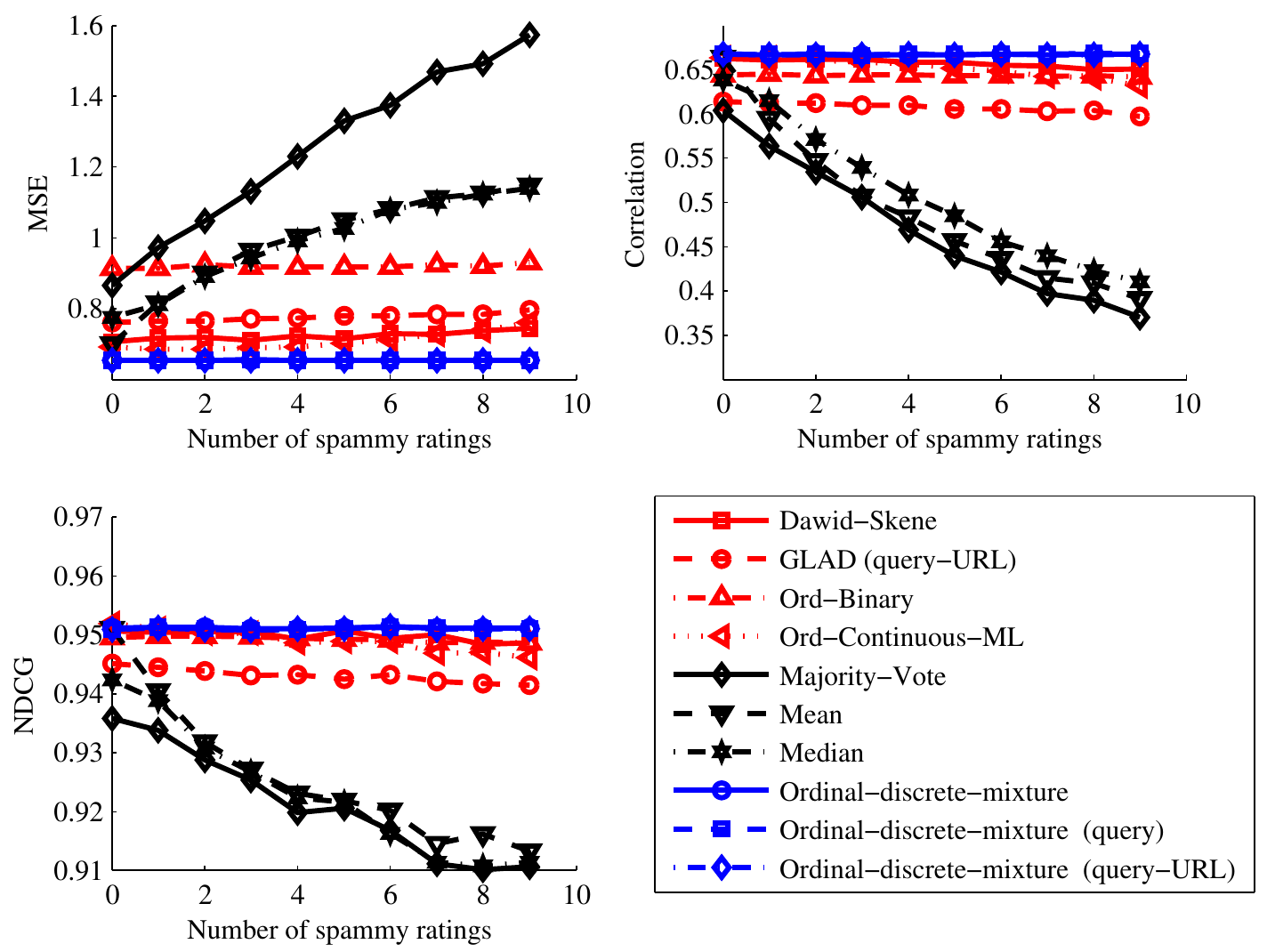}
\end{center}
\caption{Effect of spammy ratings on different methods (\yandex\ dataset): Number of additional spammy ratings per \qu\ pair verus MSE (top-left), Correlation (top-right) and NDCG (bottom-left). Variants of the proposed \modelmix\ model are shown in blue, existing state-of-the-art methods that account for annotator expertise (discussed in Section \ref{sec:relatedwork}) are shown in red and baselines that do not account for annotator expertise are shown in black. 
Baselines (mean, median, majority-vote) which do not account for annotator expertise perform significantly worse as the number of spammy ratings increases. The proposed \modelmix\ model outperforms other methods.}
\label{fig:spam}
\end{figure}

Real world crowdsourced data is often noisy and it is desirable to identify `spammy' annotators and down-weight their ratings. In this experiment, we analyze the effect of spammy ratings on the different models. For simplicity, we just present results on the \yandex\ dataset in this experiment. For every \qu\ pair in the \yandex\ dataset, we introduce additional `fake' spammy ratings drawn from an uniform distribution. We introduced fake annotators and assigned these fake ratings to fake annotators such that the average number of ratings for a fake annotator is the same as the average number of ratings for a real annotator.  We vary the number of fake ratings per \qu\ pair from 0 to 9 and repeat the previous experiment. The results are shown in Figure~\ref{fig:spam}. %
We observe that models which model annotator expertise are more resistant to spam, and simple baselines such as mean, median, and majority vote perform significantly worse in this experiment. In particular, we observe that the proposed \modelmix\ model is robust to spammy annotators and outperforms existing state-of-the-art methods discussed in Section \ref{sec:relatedwork}.

\subsection{Effect of ordinal link and mixture model}
\begin{figure}[htbp]
\begin{center}
\includegraphics[scale=1.0]{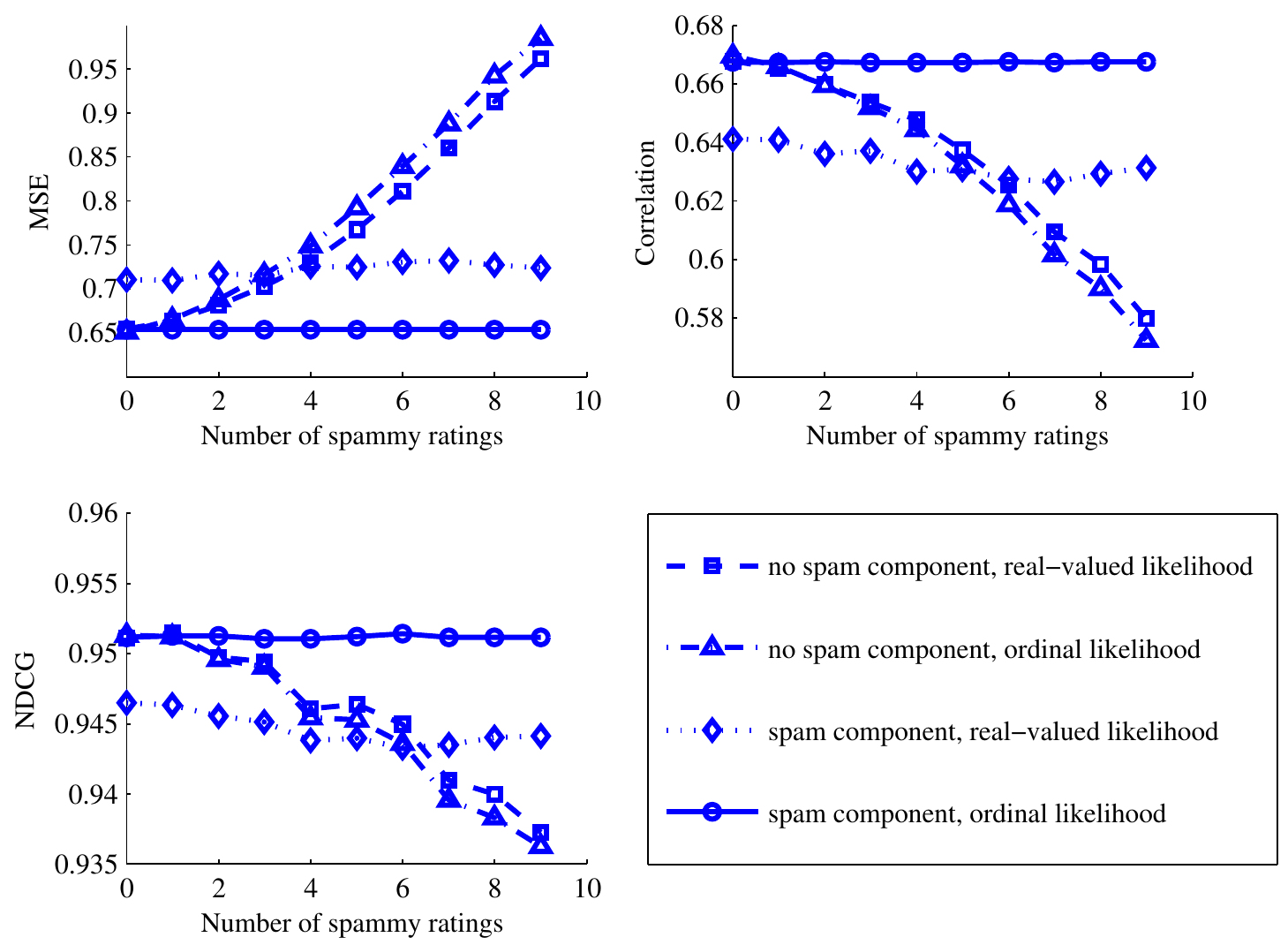}
\end{center}
\caption{Effect of spammy ratings on different variants of the \modelmix\ model (\yandex\ dataset): Number of additional spammy ratings per \qu\ pair verus MSE (top-left), Correlation (top-right) and NDCG (bottom-left). See main text for additional information.}
\label{fig:spam:ordmultinomial}
\end{figure}

In this section, we test the effect of ordinal link function and mixture model on the \modelmix\ model.
We consider four variants of the \modelmix\ model: %
whether ordinal mapping is used or not, and whether the spammy mixture component is used or not. %
Other details of the experimental setup are identical to Section \ref{sec:expt:spam}. 
To avoid clutter, we just present results for the case where instance difficulty is not modeled (other granularities for instance difficulty lead to qualitatively similar trends). 
The results are shown in Figure~\ref{fig:spam:ordmultinomial}. 
We observe that (i) the variants with the spam mixture component are robust to spammy ratings, and
(ii) amongst the variants with the spam component, the ordinal likelihood model outperforms the real-valued likelihood model. 
This experiment illustrates that both the spam mixture component and the ordinal mapping in the \modelmix\ model are necessary for good empirical performance in the presence of spammy ratings. 

\section{Conclusion} \label{sec:conclusion}%
We presented the \modelmix \ model for multi-annotator ordinal data and a scalable variational inference algorithm. The proposed model encompasses several previously proposed models. We reviewed the ordinal extension of several state-of-the-art rating models for binary/categorical/real-valued data and evaluated how well the models recover the ground truth labels. 
Our experiments on two real world datasets containing \qu\ relevance scores from AMT indicate that (i) the proposed model outperforms or performs as well as existing models in terms of MSE, correlation coefficient and NDCG and (ii) the proposed model is more resistant to spammy annotators than simple baselines which do not model annotator expertise. 
Some interesting future directions are 
 (i) joint estimation of the ground truth and optimization of the (supervised) ranking model, 
and (ii) generalizing our model to account for instance features using a Gaussian process prior in the flavor of \cite{rogers2010semi}.

\section*{Acknowledgements}
We would like to thank 
Cliff Brunk, 
Vladislav Kudelin, and 
Dmitry Pavlov of Yandex Labs for sharing the \yandex\ dataset and helpful discussions. 
We would like to thank Matthew Lease for sharing the \trec\ dataset. 
We gratefully acknowledge generous funding from the Gatsby Charitable Foundation.

\bibliography{noiseModelsTurk}
\bibliographystyle{alpha}

\appendix
\section{{Description of related work}}
\subsection{\modelmulti \ model}\label{sec:modelmulti}
In this case, the ordinal labels are treated as $K$ distinct categories. A straightforward approach is to 
model the $K\times K$ confusion matrix for each annotator  \cite{dawid1979maximum,raykar2010learning}. %
Let $\btheta=\{\bphi,\bpi\},$ where $\bpi$ is a $K$-dimensional prior such that $\pi_k=p(z_m=\vk)$ and $\bphi$ is a  $N\times K\times K$ matrix
such that $\phi_{nkj} = p(r_{nm}=\vj|z_m=\vk)$. Note that $\sum_{\jdash} \bphi_{nk\jdash} = 1 \ \forall \ n,k$, hence $\bphi$ contains $NK(K-1)$ free parameters.
The generative process can be described as follows,
\begin{align}
p(\bR|\bz,\bphi) &= \prod_{nm \in \bL} p(r_{nm}|z_m,\bphi),\label{eq:dawidskene:r}\\
p(r_{nm}|z_m,\bphi) &= \prod_{kj} \phi_{nkj}^{\indicator[r_{nm}=\vj,z_m=\vk]},\label{eq:dawidskene:noise}\\
p(\bz|\bpi) &= \prod_{m}p(z_{m}|\bpi)=\prod_{mk} \pi_k^{\indicator[z_m=\vk]}.\label{eq:dawidskene:z}
\end{align}
Note that this model doesn't account for instance difficulty. 
Inference is based on the EM algorithm, %
 treating
$\bz$ as the latent variables and $\btheta$ as the parameters. The E-step and M-step updates are given by 
\begin{align}{
\lambda_{mk} = p(z_m=\vk|\bR,\btheta)%
&\propto \pi_k \prod_{n\in \bl_m} \prod_j \phi_{nkj}^{\indicator[r_{nm}=\vj]}, \nn\\%\label{eq:lambdamk} %
\pi_k &= \frac{1}{M} \sum_m \lambda_{mk},\label{eq:dawidskene:pi}\\
\phi_{nkj} &\propto \sum_{m} \lambda_{mk}\indicator[r_{nm}=\vj] \label{eq:phinkj:update}.
}\end{align}
The above equations need to be normalized such that $\sum_{\kdash} \lambda_{m\kdash}=1$ and $\sum_{\jdash}\phi_{nk\jdash}=1$.
In our experiments, we additionally imposed a symmetric Dirichlet prior on $\phi_{nk\cdot}$ with concentration parameter $\alpha=1$. 
As before, the predicted estimate is the posterior mean, which is given by 
\begin{align}
\hat{z}_m=\E[z_{m}|\bR,\btheta] = \sum_k \lambda_{mk} \vk \label{eq:posteriormean:ordmulticlass}.
\end{align}

\subsection{\modelglad \ model}\label{sec:glad}
In this section, we describe the multi-class extension proposed in \cite{whitehill2009whose}.  Note that  \cite{whitehill2009whose} did not experimentally evaluate their multi-class extension. 
Let $\alpha_n$  denote the expertise of the $n^{th}$ annotator ($-\infty\leq\alpha_n\leq\infty$, where $\alpha_n<0$ implies adversarial annotator) 
and $\beta_m\geq0$ denote the  \textit{inverse difficulty} of the $m^{th}$ instance. The generative process is similar to the \modelmulti \ model,
except that (\ref{eq:dawidskene:noise}) is replaced by 
\begin{align}
 p(r_{nm}|z_m=\vk,\alpha_n,\beta_m)  =  \sig^{\indicator[r_{nm}=\vk]}   \Bigl(\frac{1-\sig}{K-1}\Bigr)^{\indicator[r_{nm}\neq \vk]},
\end{align}
where $\sigma(\cdot)$ denotes the sigmoid function. Comparing the \modelglad\ model with the \modelmulti\ model, we see that \modelglad
\ models just $p(r_{nm}=z_m)$ unlike the $K(K-1)$ confusion matrix in  (\ref{eq:dawidskene:noise}). However, \modelglad
\ accounts for instance difficulty. 
An alternative interpretation of the model is in terms of the log-odds ratio \cite{whitehill2009whose}: the 
logit of the probability of a correct response is bilinear in $\alpha_n$ and $\beta_m$, i.e.,
\begin{align}{
\log \frac{p(r_{nm}=z_m|\alpha_n,\beta_m)}{1-p(r_{nm}=z_m|\alpha_n,\beta_m)} = \alpha_n\beta_m.
}\end{align}
Note that higher $\alpha_n$ implies higher $p(r_{nm}=\vk)$, and as $\beta_m\rightarrow 0$ i.e., instance difficulty increases, $p(r_{nm}=z_m)\rightarrow 0.5$
instead of $\frac{1}{K}$ (corresponding to a random guess). Note 
that the model also assumes that all incorrect labels are equally likely, for example $p(r_{nm}=2|z_m=1)=p(r_{nm}=K|z_m=1)$,
which is typically not a realistic assumption. 
The E-step updates are given by:
\begin{align}{
\lambda_{mk} %
 &\propto \pi_k \prod_{n\in \bl_m}  p(r_{nm}|z_m=\vk,\alpha_n,\beta_m), 
}\end{align}
and $\blambda$ is normalized as in the previous case. The M-step updates for $\bpi$ are same as  (\ref{eq:dawidskene:pi}).
Let $\Lab$ denote the lower bound on the log likelihood. %
 We follow a co-ordinate optimization approach
 for optimizing $\Lab$ \wrt $\balpha$ and $\bbeta$.  The gradients are given by
\begin{align*}{
\frac{d\Lab}{d\alpha_n} &= \sum_k\sum_{m\in\bl_n} \lambda_{mk}\beta_m\Bigl( \indicator[r_{nm}=\vk] - \sig \Bigr),\nn\\
\frac{d\Lab}{d\beta_m} &= \sum_k \lambda_{mk}\Bigl(\sum_{n\in\bl_m} \alpha_n\bigl( \indicator[r_{nm}=\vk] - \sig \bigr)\Bigr).%
}\end{align*}
We use a conjugate gradient solver\footnote{We used Carl E. Rasmussen's \texttt{minimize.m} in our experiments. The script is available at \url{http://learning.eng.cam.ac.uk/carl/code/minimize/}.} for optimizing $\Lab$ \wrt %
$\balpha$ and $\log\bbeta$. 
 In our experiments, we additionally imposed priors on $\alpha_n \sim \Normal(1,1)$, $\log\beta_m \sim \Normal(1,1)$ as suggested by \cite{whitehill2009whose}. %
The prediction is the posterior mean and can be computed using (\ref{eq:posteriormean:ordmulticlass}).

\subsection{\modelbin \ model}\label{sec:modelbin}
A simple approach to reduce ordinal labels to binary labels is to define $K-1$ binary variables as follows \cite{frank2001simple}:
\begin{align}
\tilde{z}_{mk} = \indicator[z_m>\vk], \qquad 1 \leq k < K, \label{eq:frankhall}
\end{align} 
where the tilde indicates the $K-1$ binary variables corresponding to an ordinal variable. Similarly, let $\tilde{r}_{nmk} = \indicator[r_{nm}>\vk]$. 
Let $\tilde{z}_{m\cdot}$ and $\tilde{r}_{mk\cdot}$ denote the $K-1$ dimensional binary representations of $z_m$ and $r_{nm}$ respectively. 
Raykar et al. \cite{raykar2009supervised} suggested that it is possible to extend %
their
 two-coin binary noise  model to ordinal labels using (\ref{eq:frankhall}), but did not specify the exact noise model and explore the ordinal extension in detail. Extending the model in \cite{raykarannotation} to the case where the ground truth is ordinal, we consider the following model, 
\begin{align}{
p(r_{nm}|z_m,\btheta) &=  %
\prod_{k=1}^{K-1}  p(\tilde{r}_{nmk}|\tilde{z}_{mk},\btheta), \nn\\
p(\tilde{r}_{nmk}|\tilde{z}_{mk},\btheta) &=
 \Bigl((1-\beta_{nk})^{\tilde{r}_{nmk}} \beta_{nk}^{(1-\tilde{r}_{nmk})}  \Bigr)^{(1-\tilde{z}_{mk})}  %
 \Bigl(\alpha_{nk}^{\tilde{r}_{nmk}} (1-\alpha_{nk})^{(1-\tilde{r}_{nmk})}  \Bigr)^{\tilde{z}_{mk}}\label{eq:twocoin},
}\end{align}
where $\btheta = \{\alpha_{nk}, \beta_{nk}\}_{nk}$ and $\alpha_{nk}=p(\tilde{r}_{nmk}=1|\tilde{z}_{mk}=1)$ and  $\beta_{nk}=p(\tilde{r}_{nmk}=0|\tilde{z}_{mk}=0)$ denote the \textit{sensitivity}  and 
\textit{specificity} of the $n^{th}$ annotator for the $k^{th}$ binary variable.  Hence, we use $2(K-1)$ parameters per annotator.  For $K=3$, the confusion matrix ($\tilde{z}_{m\cdot}$ vs $\tilde{r}_{nm\cdot}$) is shown in Table \ref{tab:raykarord}. Note that the model assumes that $\tilde{r}_{nm\cdot}$ can take on $2^{K-1}$ possible values, leading to non-zero likelihood values for some invalid combinations of 
$\tilde{r}_{nm\cdot}$. However, we restrict the posterior $p(\tilde{z}_{m\cdot}|\bR)$ to $K$ values by assigning zero probability to the invalid combinations in the prior $p(\tilde{z}_{m\cdot})$. Note that this model does not account for instance difficulty.

\begin{table}[htdp]
\caption{\modelbin\ model: Confusion matrix ($K\times 2^{K-1}$) for the $n^{th}$ annotator for $K=3$. Rows indicate true labels and columns indicate observed
labels. The true  labels are assumed to lie on an ordinal scale. %
The values in parenthesis indicate the $(K-1)$ binary variables defined in (\ref{eq:frankhall}). }
\begin{center}
\resizebox{\columnwidth}{!}{
\begin{tabular}{|c|c|c|c|c|}
\hline
True vs Observed  & $1$ (00) & $2$ (10) & $3$ (11)  & Invalid (01) \\ \hline
$1$ (00) & $\beta_{n1} \beta_{n2}$ & $(1-\beta_{n1})\beta_{n2}$ & $(1-\beta_{n1})(1-\beta_{n2})$ &  $\beta_{n1}(1-\beta_{n2})$ \\ \hline
$2$ (10) & $(1-\alpha_{n1})\beta_{n2}$ & $\alpha_{n1}\beta_{n2}$ & $\alpha_{n1}(1-\beta_{n2})$ & $(1-\alpha_{n1})(1-\beta_{n2})$ \\ \hline
$3$ (11) & $(1-\alpha_{n1})(1-\alpha_{n2})$ & $\alpha_{n1}(1-\alpha_{n2})$ & $\alpha_{n1}\alpha_{n2}$ & $(1-\alpha_{n1})\alpha_{n2}$ \\ \hline
\end{tabular}
}
\end{center}
\label{tab:raykarord}
\end{table}

The EM updates can be derived as follows:
\begin{align}{
\lambda_{mk} &\propto \pi_{k} \prod_n p({r}_{nm}|{z}_{m}=\vk,\btheta), & 1\leq k \leq K, \nn\\
\gamma_{m\kdash} &= \sum_k\lambda_{mk}\indicator[\tilde{z}_{m\kdash}],  & 1\leq\kdash<K,\nn\\
\alpha_{n\kdash} &= \frac{\sum_m \gamma_{m\kdash}\tilde{r}_{nm\kdash}}{\sum_m \gamma_{m\kdash}}, & \nn\\
\beta_{n\kdash} &= \frac{\sum_m (1-\gamma_{m\kdash})(1-\tilde{r}_{nm\kdash})}{\sum_m (1-\gamma_{m\kdash})}. & 
}\end{align}
The prediction is the posterior mean and can be computed using (\ref{eq:posteriormean:ordmulticlass}).

\end{document}

%% file: macros.tex
\newcommand{\real}{\mathrm{I\kern-0.175em R}}
\def\freeEn{\mathcal{F}}

\def\entropy{\mathrm{H}}

\newcommand{\expect}{\mathbb{E}}

\newcommand{\KL}[2]{\mathcal{KL}\left(#1||#2\right)}

\newcommand{\distrib}[2]{\mathcal{#1}\mathit{#2}}

\newcommand{\bernoulli}{\distrib{B}{e}}

\def\bbeta{\svct{\beta}}

\newcommand{\E}{\mathbb E}

\def\nn{\nonumber}

\def\balpha{\bm{\alpha}}
\def\bbeta{\bm{\beta}}
\def\bdelta{\bm{\delta}}
\def\blambda{\bm{\lambda}}

\def\bphi{\bm{\phi}}
\def\bpi{\bm{\pi}}

\def\btau{\bm{\tau}}
\def\btheta{\bm{\theta}}

\def\bv{\bm{v}}

\def\bz{\bm{z}}

\def\bR{\bm{R}}
\def\bL{\bm{L}}
\def\bX{\bm{X}}
\def\bY{\bm{Y}}

\def\Normal{\mathcal{N}}

\def\cdash{c^{\prime}}

\def\jdash{j^{\prime}}
\def\kdash{k^{\prime}}

\def\indicator{\mathds{1}}

\def\Var{\mathop{\rm Var}}

%% file: noiseModelsTurk.bbl
\newcommand{\etalchar}[1]{$^{#1}$}
\begin{thebibliography}{WRW{\etalchar{+}}09}

\bibitem[Bea03]{Beal03}
M.J. Beal.
\newblock {\em Variational Algorithms for Approximate {B}ayesian Inference}.
\newblock PhD thesis, Gatsby Unit, University College London, 2003.

\bibitem[BGMG12]{bachrach2012grade}
Y.~Bachrach, T.~Graepel, T.~Minka, and J.~Guiver.
\newblock {How To Grade a Test Without Knowing the Answers---A {B}ayesian
  Graphical Model for Adaptive Crowdsourcing and Aptitude Testing}.
\newblock {\em ICML}, 2012.

\bibitem[BMLS10]{Buckley10-notebook}
C.~Buckley, M.~Matthew~Lease, and M.D. Smucker.
\newblock {Overview of the TREC 2010 Relevance Feedback Track (Notebook)}.
\newblock In {\em {The Nineteenth Text Retrieval Conference (TREC) Notebook}},
  2010.

\bibitem[Car08]{carpenter2008multilevel}
B.~Carpenter.
\newblock Multilevel {B}ayesian models of categorical data annotation.
\newblock {\em Unpublished manuscript}, 2008.

\bibitem[CG05]{chu2005gaussian}
W.~Chu and Z.~Ghahramani.
\newblock Gaussian processes for ordinal regression.
\newblock {\em JMLR}, 6:1--48, 2005.

\bibitem[DS79]{dawid1979maximum}
A.P. Dawid and A.M. Skene.
\newblock Maximum likelihood estimation of observer error-rates using the {EM}
  algorithm.
\newblock {\em Applied Statistics}, pages 20--28, 1979.

\bibitem[FH01]{frank2001simple}
E.~Frank and M.~Hall.
\newblock A simple approach to ordinal classification.
\newblock {\em ECML 2001}, pages 145--156, 2001.

\bibitem[GR06]{girolami2006variational}
M.~Girolami and S.~Rogers.
\newblock Variational bayesian multinomial probit regression with gaussian
  process priors.
\newblock {\em Neural Computation}, 18(8):1790--1817, 2006.

\bibitem[How06]{howe2006rise}
J.~Howe.
\newblock The rise of crowdsourcing.
\newblock {\em Wired magazine}, 14(6):1--4, 2006.

\bibitem[KOS11]{karger2011iterative}
D.R. Karger, S.~Oh, and D.~Shah.
\newblock Iterative learning for reliable crowdsourcing systems.
\newblock {\em NIPS}, 2011.

\bibitem[Liu09]{liu2009learning}
T.Y. Liu.
\newblock Learning to rank for information retrieval.
\newblock {\em Foundations and Trends in Information Retrieval}, 3(3):225--331,
  2009.

\bibitem[RGP10]{rogers2010semi}
S.~Rogers, M.~Girolami, and T.~Polajnar.
\newblock Semi-parametric analysis of multi-rater data.
\newblock {\em Statistics and Computing}, 20(3):317--334, 2010.

\bibitem[RY11]{raykarannotation}
V.C. Raykar and S.~Yu.
\newblock Annotation models for crowdsourced ordinal data.
\newblock {\em NIPS workshop on Computational Social Science and the Wisdom of
  Crowds}, 2011.

\bibitem[RYZ{\etalchar{+}}09]{raykar2009supervised}
V.C. Raykar, S.~Yu, L.H. Zhao, A.~Jerebko, C.~Florin, G.H. Valadez, L.~Bogoni,
  and L.~Moy.
\newblock Supervised learning from multiple experts: Whom to trust when
  everyone lies a bit.
\newblock {\em ICML}, 2009.

\bibitem[RYZ{\etalchar{+}}10]{raykar2010learning}
V.C. Raykar, S.~Yu, L.H. Zhao, G.H. Valadez, C.~Florin, L.~Bogoni, and L.~Moy.
\newblock Learning from crowds.
\newblock {\em JMLR}, 2010.

\bibitem[SF08]{sorokin2008utility}
A.~Sorokin and D.~Forsyth.
\newblock Utility data annotation with amazon mechanical turk.
\newblock {\em Computer Vision and Pattern Recognition Workshops}, 2008.

\bibitem[SFB{\etalchar{+}}95]{smyth1995inferring}
P.~Smyth, U.~Fayyad, M.~Burl, P.~Perona, and P.~Baldi.
\newblock Inferring ground truth from subjective labelling of venus images.
\newblock {\em NIPS}, 1995.

\bibitem[SOJN08]{snow2008cheap}
R.~Snow, B.~O'Connor, D.~Jurafsky, and A.Y. Ng.
\newblock Cheap and fast---but is it good?: Evaluating non-expert annotations
  for natural language tasks.
\newblock {\em EMNLP}, 2008.

\bibitem[WBBP10]{welinder2010multidimensional}
P.~Welinder, S.~Branson, S.~Belongie, and P.~Perona.
\newblock The multidimensional wisdom of crowds.
\newblock {\em NIPS}, 2010.

\bibitem[WRW{\etalchar{+}}09]{whitehill2009whose}
J.~Whitehill, P.~Ruvolo, T.~Wu, J.~Bergsma, and J.~Movellan.
\newblock {Whose vote should count more: Optimal integration of labels from
  labelers of unknown expertise}.
\newblock {\em NIPS}, 2009.

\bibitem[YRF{\etalchar{+}}10]{yan2010modeling}
Y.~Yan, R.~Rosales, G.~Fung, M.~Schmidt, G.~Hermosillo, L.~Bogoni, L.~Moy, and
  J.~Dy.
\newblock Modeling annotator expertise: Learning when everybody knows a bit of
  something.
\newblock {\em AISTATS}, 2010.

\end{thebibliography}
